\documentclass[a4paper,10pt]{amsart}
\pdfoutput=1
\usepackage{amsmath, amssymb, amsthm, amsfonts, amscd, epsfig, graphicx, color}

\theoremstyle{plain}
\newtheorem{theo}{Theorem}[section]

\newtheorem{rem}[theo]{Remark}

\numberwithin{equation}{section}

\newcommand{\R}{\mathbb R}

\newcommand{\punto}{\xi}
\newcommand{\puntoprimo}{\xi '}

\title[The constitution of visual perceptual units] { The constitution of visual perceptual units in the functional architecture of V1 }

\author{Alessandro Sarti and Giovanna Citti}

\begin{document}
\maketitle

\section{Introduction}

One of the major challenges in neurobiology is understanding the relationship between
spatially structured activity states and the underlying neural circuitry that supports them.

From the geometrical point of view the first 
accurate models of the functional architecture of the primary visual cortex (V1) is due to Hubel 
and Wiesel \cite{Hubel1977} (see \cite{Hubel} for a review of their work). Hubel and Wiesel 
discovered that for every point $(x,y)$ of the retinal plane there is an entire set of cells, each one sensitive to a particular instance of a specific feature of the image: position, orientation, scale, color, curvature, velocity, stereo. They called this structure hypercolumnar organization.  Horizontal connectivity is responsible for the cortico-cortical propagation of the neural activity between hypercolumns.
Further insights on the structure of the connectivity and the spatial arrangements of cells were provided by  \cite{Blasdel}, \cite{Bonhoffeer}, \cite{Bosking}. 
The association fields of Field, Hayes and Hess \cite{Field1993}, discovered on a purely psycho-physical basis, have been proposed as a phenomenological counterpart of the cortical-cortical connectivity.
Geometric frameworks for the description of the functional architecture of V1
were proposed by W.C. Hoffmann in  \cite{Hoffman}, Petitot and Tondut \cite{Petitot1999}, Bressloff an Cowan \cite{BressloffCowan}, Citti and Sarti \cite{CS}, Zucker \cite{Zucker2006},  Sarti, Citti, Petitot \cite{SCP}. Application to image processing can be found in \cite{Duits2010a}, \cite{Duits2010b}, \cite{Duits2011}.

From the dynamical point of veiw the first neural field models of the cortical activity are due to Wilson and  Cowan \cite{WilsonCowan1972, WilsonCowan1973} and Amari \cite{Amari}, and  are  expressed in terms of integro-differential equations. 
Extensions of the models have been provided by Ermentrout and Cowan \cite{Ermentrout1979, Ermentrout1980}. 
These mean field equations  describe the activity on a 2D plane and formally 
express the interaction between cells through as a convolution kernel.
 Bressloff and Cowan \cite{BressloffCowan, BressloffCowan2002} proposed 
new models taking into account the high dimensional cortical structure, with  orientation and scale as  eingrafted variables. 
In their models the connectivity kernel satisfies the symmetry properties of the cortical space, namely $SE(2)$ for rotation and translation and the affine group for scale, rotation and translation.
In absence of the external input these models successfully account for hallucination patterns. 
More recently , Faugeras \cite{Faugeras}, Faye and Faugeras in \cite{FayeFaugeras} and Chossat, Faye, Faugeras \cite{ChossatFayeFaugeras} modified the model in order to take into account delay and the tensorial structure of the cortex.

Scope of this paper is to consider a mean field neural model which takes  into account the neurogeometry of the cortex introduced in \cite{CS} as well as the presence of a visual input. 
It is known that when stationary solutions of the equation become marginally stable, eigenmodes of the linearized operator can become stable.  In absence of a visual input the raising eigenmodes lead to the hallucination patterns proposed by 
Bressloff and Cowan  \cite{BressloffCowan, BressloffCowan2002}.
The main result of our study consists in showing that in presence of a visual input, these eigenmodes corresponds to perceptual units. While in the case of hallucinations the emergence of eigenmodes is due to the use of drugs, 
in the case of perceptual units it is due to physiological 
variations of parameters during the perception process. 
The whole process can be interpreted as a problem of data segregation and partitioning, 
strongly related to the most recent results of dimensionality reduction.
In particular our model can justify on biological basis, the results of  \cite{Perona, ShiMalik, Weiss, 
CoifmanLafon, Coifman}, who directly faced the problem of perceptual 
grouping in the description of a scene 
 by means of a kernel PCA on an affinity matrix. 

The paper starts with briefly recalling some results about the neurogeometry of the primary visual cortex (section 2).  The cortico-cortical interaction between simple cells is represented by the fundamental solution of a Fokker Planck equation,  following \cite{SCS} and \cite{BCCS}. In section 3 the classical mean field model of Ermentraut and Cowan is adapted to the $SE(2)$ cortical symmetry group with the previously computed connectivity kernel. Stationary solutions are studied and a stability analysis is performed, varying a suitable physiological parameter. In the classical papers \cite{BressloffCowan, BressloffCowan2002} the variability of this parameter was due to the presence of drugs. On the contrary in our model, the variability of the same parameter is due to the physiological variability of 
the transfer function in different neural populations. In addition,  
the geometry of the problem depends both on the invariance of $SE(2)$ and the presence of the input. 
In section 4 the mean field equation is discretized and the connectivity kernel reduced to a matrix induced by the neurogeometry of the cortex as well as by the visual input. 
Marginally stable solutions are computed as eigenvectors of this matrix, 
and we show that they represent perceptual units present in the image. 
The result is strictly related to the dimensionality reduction and clustering problems of  \cite{Perona}, 
and the connectivity matrix can be interpreted as an affinity matrix. 
Finally in section 5 we present numerical simulation results.

\section{The functional geometry of V1}\label{functionalgeometry}

In this section we briefly recall the structure of the functional geometry of the visual cortex. As proved by 
Hubel and Wiesel \cite{Hubel1977} the  visual cortex is organized in hypercolumns of simple cells sensitive to the position $(x, y)$ and eingrafted variables, which describe different properties of the stimulus: orientation, 
curvature, speed, velocity, scale, disparity. We will describe in detail the structure of the family of simple cells, sensible to 
position and orientation.

\subsection{The $SE(2)$ symmetry of the visual cortex}

  Many authors \cite{Petitot1999, CS, Zucker2006} represented the hypercolumnar organization as a 3-dimensional space with coordinates $(x, y, \theta)$ where each point corresponds to a specific population of cells sensitive to a stimulus positioned in $(x, y)$ and with orientation $\theta$. This leads to the description of the visual cortex in the  special Euclidean group  group $SE(2) \approx \R^2 \times S^1$.
This  is composed by the semi-direct product of  the group of translations of the plane $\mathbb{R}^2$ with the rotations and reflections group of the plane   $O(2)$.
The action $\cdot$ of the Euclidean group
on every element $(x,y,\theta)$ of $\mathbb{R}^2\times\mathbf{S}^1$  is generated by:
\begin{itemize}
 \item $(x',y')\cdot(x,y,\theta)=(x+x', y+y',\theta)\textrm{,}\quad (x',y')\in\mathbb{R}^2 \textrm{is the translation vector;}$
 \item $\theta'\cdot(x,y,\theta)=\left(R_{\theta'}(x,y),\theta'+\theta\right)\textrm{,}\quad\theta'\in\mathbf{S}^1\textrm{,}$\\
       $R_{\theta}\textrm{ is the rotation matrix of an angle }\theta'\textrm{;}$
 \item $k\cdot(x,y,\theta)=(x,-y,-\theta)\textrm{,}\\ \text{ if }k:(x,y)\mapsto(x,-y)\textrm{ is the reflection transformation}\textrm{.}$
\end{itemize}

\begin{figure}
\centering
\includegraphics[height=6cm]{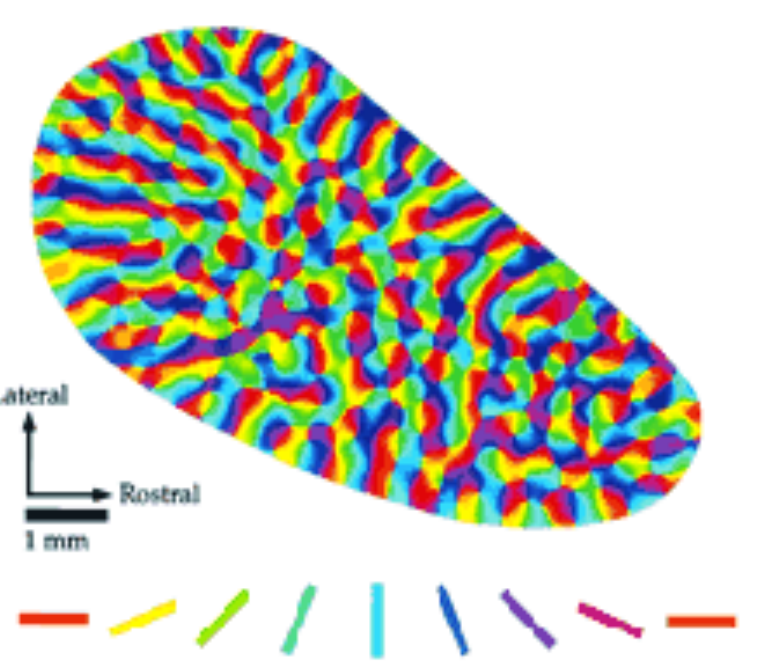}
\caption{The pinwheel structure of the primary visual cortex measured by in vivo optical imaging taken from \cite{Bosking}. Orientation maps are coded with the colorbar on the right.}\label{fig3pin}
\end{figure}

Consequently we will also denote 
$$(x',y', \theta') \cdot (x,y,\theta) = \left(R_{\theta'}(x,y) + (x', y'),\theta'+\theta\right),$$
which defines the composition law in the Euclidean group. 
 
This 3-dimensional group is implemented in the two dimensional 
layer of the visual cortex where the position-orientation features are coded 
simultaneously in the pinwheel structure (see Figure \ref{fig3pin}).

\subsection{The output of simple cells as cortical lifting of the visual stimulus}
The receptive profile of a simple cell has been modelled as a Gabor filter  
or in 
terms of derivatives of a Gaussian function (\cite{Daugmann}). 
The whole set of simple cells $\psi_{(x,y,\theta)}$  can be obtained by rotation and translation from the mother filter $\psi_{(0,0,0)}$, which 
amount to say that for every $(x,y,\theta)$ the cell at position $(x,y)$ sensible to the orientation $\theta$ can be represented as
\begin{equation}\label{actioncells}\psi_{(x,y,\theta)} (x',y')=
\psi_{(0,0,0)}\big(R_\theta(x'-x,y'-y)\big).\end{equation}

 The response of simple cells to a visual stimulus $I(x,y)$ can be obtained as an integral of the RP with the image $I$:
\begin{equation}\label{input} h(x,y,\theta)= \int \psi_{(x,y,\theta)}(x',y') I(x',y') dx'dy'.\end{equation}
Since the output depends on the $(x,y,\theta)$ variables of the Euclidean motion group, the action of the cells is modeled as a lifting process of the retinal 2D image $I(x,y)$ to a function $h(x,y,\theta)$ defined on the Lie group manifold.

\subsection{Geometry of the horizontal connectivity}

\begin{figure}
\centering
\includegraphics[width=5.5cm]{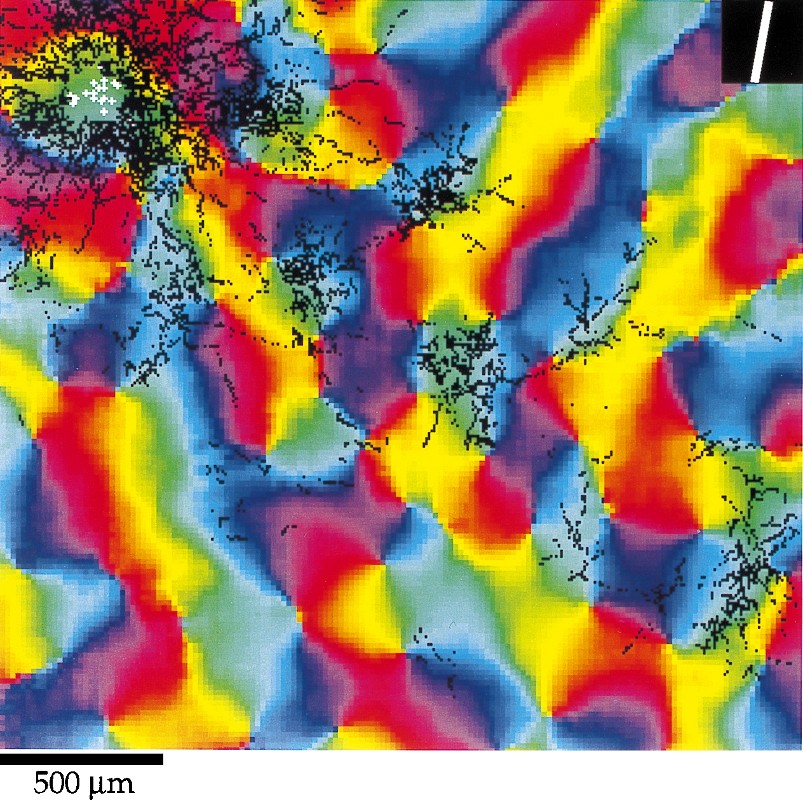}
\caption{Cortico-cortical connectivity measured by Bosking in \cite{Bosking}. The tracer is propagated through the lateral connections to points in black. These locations are plotted together with the orientation maps.}
\label{tracer}
\end{figure}

Hypercolumns are connected by means of the so called horizontal or cortico-cortical connectivity. 
Experimental measures of this connectivity have been obtained by Bosking in \cite{Bosking} by injecting a chemical tracer (biocytin) and observing its propagation in the cortical layer (see  Figure \ref{tracer}).

In \cite{CS}  the connectivity between hypercolumns has been described in terms of the Lie algebra of $SE(2)$, and the following vector fields have been chosen as its generators at a general point $(x,y,\theta)$:
\begin{equation}\label{contact}
\vec{X_1} = (\cos\theta, \sin\theta, 0), \quad \vec{X_2} = (0, 0, 1) \ .
\end{equation}

The points of the structure are connected by integral curves of these two vector fields:
$$c:\R\rightarrow SE(2), \quad c(s)=(x(s),y(s),\theta(s))$$ 
such that 
\begin{equation}
c'(s) =(k_1(s)  \vec{X_1} + k_2(s) \vec{X_2})(c(s)), \,\,\, c(0) = 0.
\label{curvhor}\end{equation}

The length of these curves is computed
as $$l(c) = \int ||c'(s)||ds$$
where different norms can be defined, with different choice of exponents for the components of $k$:
\begin{equation}\label{norm}||c'(s)|| =(|k_1(s)|^{n_1} +| k_2(s)|^{n_2})^{1/2}.\end{equation}
The choice $n_1=1$, $n_2=2$ is compatible with the Fokker Planck model proposed by Mumford in \cite{Mumford} and Williams and Jacobs in \cite{Williams}. 
With the choice of exponents $n_1 = n_2=2$ we obtain a different norm, associated to the subriemannian Laplacian. 
Since any couple of points $(x,y,\theta)$ $(x',y',\theta')$ can be connected by 
such integral curves, a Carnot Carath\'eodory distance can be defined:
\begin{equation}\label{dCC}\it{d_c}\Big((x,y,\theta), (x',y',\theta')\Big)= \inf\{l(c): c \text{ connects  } (x,y,\theta) \text{ and } (x',y',\theta')\}\end{equation}
(see for example  Nagel Stein Wainger \cite{NSW} or Mongomery \cite{Mongomery}). 
Sanguinetti Citti Sarti showed that this distance associated with the Fokker Planck choice fits very well the statistics of co-occurrence of edges 
in natural images \cite{SCS}.  See also \cite{Sachov} for some results on geodesics in this setting. 

\subsection{The connectivity kernel as fundamental solution}
The cortical connectivity can be modeled with the 
stochastic counterpart of the curves defined in (\ref{curvhor}): 
\begin{equation}\label{stoc} (x', y',
\theta') = (\cos (\theta),\sin(\theta) ,N (0, \sigma^2)) = \vec{X}_1 + N
(0, \sigma^2)\vec{X}_2
\end{equation}
 where $N(0, \sigma^2)$ is a normally distributed variable with zero
mean and variance equal to $\sigma^2$. 
This approach, first introduced by Mumford
in \cite{Mumford}, has  been further discussed by August-Zucker
\cite{AugustZucker2000, AugustZucker2003}, Williams-Jacobs \cite{Williams}, and
Sanguinetti-Citti-Sarti \cite{SCS}, and we shortly recall it here. 

Let's denote $u$ the transition probability that the stochastic solution starting from the point
$(x', y')$ with orientation $\theta'$ at the initial time reaches the point 
$(x, y)$ with orientation $\theta$ at the time $s$. This probability density satisfies a
deterministic equation known in literature as the Kolmogorov Forward
Equation or Fokker-Planck equation (FP):
\begin{equation}\label{(FP)}
\partial_t v=X_{1} v +
\sigma^2X_{22}v
\end{equation}
 where $X_1$ is the directional derivative $cos(\theta) \partial_x + sin(\theta) \partial_y$ and $X_2= \partial_\theta$, while  $X_{22}= \partial_{\theta\theta}$ is the second order derivative.

This
equation has been largely used in computer vision and applied to
perceptual completion related problems. It was first used by
Williams and Jacobs \cite{Williams} to compute stochastic
completion field, by   August and Zucker \cite{AugustZucker2000, AugustZucker2003} to define the curve indicator
random field, and more recently by R. Duits and Franken in
\cite{Duits2010a, Duits2010b}  to perform
contour completion, de-noising and contour enhancement.
Its stationary counterpart was proposed in \cite{SCS08}
to model the probability of co-occurence of contours in natural
images:
\begin{equation}\label{FPoperator}FP =X_1  + \sigma^2 X_{22} 
\end{equation} 
This operator has a nonnegative fundamental solution $\Gamma$ satisfying: 
\begin{equation}\label{gonzalo}X_1 \Gamma((x, y, \theta),(x', y', \theta')
) + \sigma^2 X_{22} \Gamma((x, y, \theta),(x', y', \theta')) = \delta(x, y,
\theta),
\end{equation}
The kernel is strongly biased in direction $X_1$ and not symmetric. Its symmetrization can be 
obtained as: 
\begin{equation}\label{omegakernel}\omega((x,y,\theta),(x',y',\theta')) =\frac{1}{2}\Big(\Gamma((x,y,\theta),(x',y',\theta')) + \Gamma((x',y',\theta'),(x,y,\theta))\Big).\end{equation}

We explicitly recall that the general 
 results of \cite{RS} and \cite{NSW} provide a local estimate of the fundamental solution 
(and consequently also of $\omega$) in terms of the distance defined in the previous section. 
The choice of exponents $n_1=2$, $n_2=1$ we made in equation (\ref{norm}) 
is compatible with the fact that in this equation we have a first derivative 
in direction $X_1$ and a second one in direction $X_2$. Indeed 
the kernel $\omega$ is estimated in terms of the distance $\it{d_c},$ as follows:
 \begin{equation}\label{omegadist}\omega ((x,y,\theta), (x',y',\theta') )\simeq e^{-\it{d_c}^2((x,y,\theta), (x',y',\theta')) },\end{equation}

Let's also recall that this model closely matches the statistical distribution of edge co-occurence in natural images as obtained in \cite{SCS08}. This argument strongly suggests that horizontal connectivity modelled by the neurogeometry is deeply shaped by the statistical distributions of features in the environment and that the very origin of neurogeometry has to be discovered in the interaction between the embodied subject and the world.

\section{Mean field equation in the cortical space}

The evolution of a state of a population of cells
has been modelled by Wilson and Cowan in  \cite{WilsonCowan1972}, \cite{WilsonCowan1973},  by Ermentrout and Cowan  \cite{Ermentrout1980}, and subsequently by  Bressloff and Cowan in \cite{BressloffCowan}. Recent results are due to Faye and Faugeras \cite{FayeFaugeras} and Chossat, Faye, Faugeras \cite{ChossatFayeFaugeras}. 
The Ermentraut Cowan mean field equation rewritten in the cortical space reads

\begin{equation} \label{eq:2}
\frac{\textrm{d}a(\punto,t)}{\textrm{d}t}=-\alpha a(\punto,t)+\sigma\Big(\int \mu \omega(\punto, \punto') a(\punto', t) d\punto' + h(\punto, t) \Big)\quad \textrm{ in } \mathcal{M}
\end{equation}
where $\xi=(x,y,\theta) $ is a point of the cortical space $\mathcal{M}$, the coefficient $\alpha$ represents the
decay of activity, $h$ is the feedforward input which coincides with the response of the simple cells in presence of a visual stimulus described by (\ref{input}). 

\begin{figure}
\centering
\includegraphics[width=7.5cm]{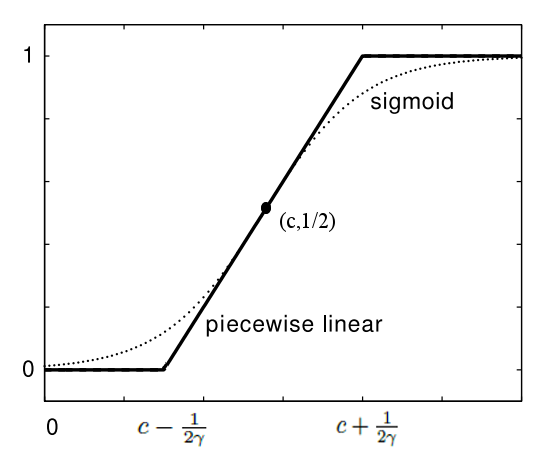}
\caption{The piecewise linear transfer function, compared with the classical sigmoid.}
\label{sigmono}
\end{figure}

The function $\sigma$ is the transfer function of the population, and has a piecewise linear behavior, as proposed in \cite{Kilpatrick} (see Figure \ref{sigmono}).

\begin{equation}\label{sigmoidal} \sigma(s) = 
\left\{\begin{array}{rcl}
0, & s\in ]-\infty, c-\frac{1}{2\gamma}[\\
\\
\gamma(s-c)+\frac{1}{2}, & s\in [c-\frac{1}{2\gamma}, c+\frac{1}{2\gamma}]\\
\\
1, & s\in ]c+\frac{1}{2\gamma}, +\infty[
\end{array}\right.
,\end{equation}
where $\gamma$ is a real number, which represents the slope of the linear regime 
and $c$ is the half height threshold.

The kernel  $\mu\omega(\punto,\puntoprimo)$ is the contribution of cortico-cortical connectivity introduced in (\ref{omegakernel}). It is compatible with the model of Bressloff and Cowan who only assumed that $\omega$ is invariant with respect to 
rotation and translations. The parameter $\mu$ is a coefficient 
of short term synaptic facilitation and generally increasing during the perceptual 
process.

We also outline the following existence result: 
\begin{rem}\label{rem1} {\bf{Existence of the solution.}} The solution is defined for all times and satisfies 
$$|a(\xi, t)|\leq \frac{1}{\alpha} \text{ for all } \xi, \in M, t >0.$$
See for example \cite{Faugeras3}.
\end{rem}

\subsection{Restriction to the domain defined by the external input}

The main novelty of our model is to split the cortical domain $M$ in a subdomain 
$\Omega$ characterized by the presence of the input, and the complementary set. 
We will show in the following that 
under suitable assumptions the activity in this complementary set will be 
negligible and the domain of equation (\ref{eq:2}) reduces to $\Omega$.

By simplicity we will assume that $h$ can attain only two values: $0$ and $c$, 
and we call $\Omega$ the set of points in the visual cortex activated by the presence of an input
\begin{equation}\label{hdomain}\Omega=\{\xi: h(\xi)=c \}.\end{equation}
We require that $\mu\omega$ satisfies an assumption of weak connectivity, 
which means that when the activity is around the points $0$ and $c$, 
the dynamics does not change regime due to the connectivity contribution. 

\begin{rem}
Formally we will require that the integral of $\mu\omega$ is sufficiently small to satisfy:
\begin{equation}\label{weak}\int_M \mu \omega(\xi, \xi') d\xi' \leq \alpha \min\Big( \frac{1}{2\gamma}, c - \frac{1}{2\gamma}\Big).\end{equation}

Under this assumption, if the activity $a$ is identically $0$ at the initial time, 
then the activity remains identically $0$ outside $\Omega$ for all $t>t_0$: 
$$a(\xi, t)=0  \text{ for } \xi\in M\backslash\Omega.$$ 
On the other hand on the set $\Omega$ the argument of $\sigma$ always remains in the linear regime for all $t>t_0$:
\begin{equation}\int \mu \omega(\xi, \xi') a(\xi')d\xi' + c\in [c- \frac{1}{2\gamma}, c+ \frac{1}{2\gamma}], \quad \text{ for } \xi\in \Omega.\end{equation}
\begin{proof}

Let us choose $\xi$ in $M\backslash \Omega$. Using the boundness of 
$a$ asserted in Remark \ref{rem1}, and the assumption of weak connectivity (\ref{weak}) 
on $\omega$ we get
\begin{equation}\label{omegastima}\int \mu\omega(\xi, \xi') a(\xi') d\xi' \leq \frac{\alpha\max(a)}{2\gamma} \leq \frac{1}{2\gamma}.\end{equation}
 It follows that 
$$\sigma \Big(\int \mu\omega(\xi, \xi') a(\xi') d\xi' \Big) =0,$$
if $\xi\in M\backslash\Omega$. Inserting this in the right hand side of equation (\ref{eq:2})
$$\frac{da}{dt} (e^{\alpha t} a(\xi, t))  = e^{\alpha t} a'(\xi, t) + \alpha e^{\alpha t} a(\xi, t) $$$$=e^{\alpha t} \sigma \Big(\int \mu\omega(\xi, \xi') a(\xi') d\xi'\Big)=0,$$
This implies that 
$$e^{\alpha t} a(\xi, t)$$ is constant, and since it vanishes for $t=t_0$, it is identically $0$ for all $t>t_0$. 
From (\ref{omegastima}) it also follows that 
$$\int \mu \omega(\xi, \xi') a(\xi')d\xi' + c \leq c + \frac{1}{2\gamma}$$
and 
$$\int \mu \omega(\xi, \xi') a(\xi')d\xi' + c \geq c - \frac{1}{2\gamma}.$$

\end{proof}

\end{rem} 

 Hence the mean field activity equation reduces to 
\begin{equation}\label{eqrem}
\frac{\textrm{d}a(\punto,t)}{\textrm{d}t}=-\alpha a(\punto,t)+\gamma \Big(\int \mu\omega(\xi, \xi') a(\xi', t) d\xi' + c \Big)\quad \textrm{ in } \Omega. 
\end{equation}

Note that the equation (\ref{eqmeanfield}) is similar to the one in Bresslof Cowan model, but the Bresslof Cowan model  is defined in the whole cortical space, while equation (\ref{eqmeanfield}) is defined on the domain $\Omega$.

\subsection{Stability analysis}

The stationary states $a_1$ of equation (\ref{eqrem}) satisfy 

\begin{equation}\label{stazionario}-\alpha a_1(\punto)+\gamma\Big(\int \mu\omega(\xi, \xi')  a_1(\xi') d\xi' + c \Big)=0\quad \textrm{ in } \Omega\end{equation}
and have been studied by \cite{Faugeras3}.

In order to study their stability we need to study small perturbation around the stationary 
state. Hence we will call $u= a-a_1$ the perturbation,  
and obtain the equation satisfied by $u$ subtracting the equations for $a$ and $a_1$:

$$\frac{\textrm{d}(a-a_1)(\punto,t)}{\textrm{d}t}=-\alpha (a-a_1)(\punto,t)+\gamma\Big(\int \mu\omega(\xi, \xi') ( a - a_1 ) (\xi', t) d\xi' \Big)$$

in $\Omega$. Note that the function $u$ is a solution of the homogeneous equation associated to 
(\ref{eqrem}):

\begin{equation} \label{eqmeanfield}
\frac{\textrm{d}u(\punto,t)}{\textrm{d}t}=-\alpha u(\punto,t)+\gamma\Big(\int \mu\omega(\xi, \xi') u(\xi') d\xi' \Big)\quad \textrm{ in } \Omega
\end{equation}

The stability of the solution of this linear equation can be studied by means of the eingenvalues of the 
associated linear operator:

\begin{equation}Lu = - \alpha u + \mu\gamma\int \omega(\xi, \xi') u(\xi') d\xi'
 =\lambda u.\label{Leig}\end{equation}

Let us note that the parameter $\mu$ increases since it is a short term synaptic facilitation. For this reason we now study this eigenvalue problem by varying $\mu$. The system will be stable if $\lambda$ 
is negative. This condition depends on the value of $\mu$ and on the eigenvalues of the convolution operator with $\mu\omega$. Indeed condition \ref{Leig}
is equivalent to 
$$ \int  \omega(\xi, \xi') u(\xi') d\xi' = \frac{1}{\gamma\mu}(\lambda + \alpha) u.$$
and implies 
$$\frac{\lambda + \alpha}{\gamma\mu} = \tilde \lambda$$
for an eigenvalue $\tilde\lambda$ of $\omega$. 
Imposing that $\lambda$ is negative we get: 
$$\lambda = - \alpha  + \mu\gamma \tilde \lambda<0$$
Hence 
$$ \mu <   \frac{\alpha}{\gamma\tilde \lambda} $$
for every eigenvalue $\tilde \lambda$ of $\omega$. Remember that the operator associated to $\omega$ has a sequence $\tilde \lambda_k$ of eigenvalues. This is satisfied if 
$$ \mu <   \frac{\alpha}{\gamma\tilde \lambda_1},$$ 
for the largest eigenvalue $\tilde \lambda_1$. 
The uniform solution becomes marginally stable 
when $\mu$ increases beyond the critical value $ \frac{\alpha}{\gamma\tilde \lambda_1}$. 
due the excitation of the linear eigenfunctions, solutions of 
\begin{equation}\label{eigenomega} \int  \omega(\xi, \xi') u(\xi') d\xi' = \tilde \lambda_k u.\end{equation}
 The saturating
nonlinearities of the system can stabilize the growing pattern
of activity.

\section{Patterns of activity and spectral clustering}

\subsection{The discrete mean field equation}
Due to the discrete structure of the cortex, 
the input configurations are 
constituted by a finite number $N$ of position-orientation elements, 
with coordinates $\xi_i=(x_i, y_i, \theta_i)$. 
On these points the input $h$ takes the value $c$. 
As a consequence, the set $\Omega$, defined in (\ref{hdomain}) is discretized,
and becomes
\begin{equation}\label{omegadiscrete}\Omega_d =\{\xi_i: h(\xi_i)=c\}.\end{equation}
Analogously the main field equation (\ref{eqmeanfield}) reduces to:
\begin{equation} \label{eqdiscrete}
\frac{\textrm{d}a(\punto_i,t)}{\textrm{d}t}=-\alpha a(\punto_i,t)+
\gamma\mu\sum_{j=1}^N\omega(\punto_i,\punto_j)a (\punto_j, t) \quad \textrm{ in } \Omega_d. 
\end{equation}

The model of \cite{BressloffCowan} has been developed in the 
whole cortical space without an input and the activity patterns have the symmetry of $SE(2)$. 
Here the symmetry is lost due to the presence of the input, hence the activity patterns 
inherit geometric properties of the domain $\Omega_d$. The eigenmodes 
will be defined precisely on that geometry. 

In particular the kernel $\omega$ is reduced to a matrix $A$, whose entries $i,j$ are: 
\begin{equation}\label{affinemat}A_{ij} =  \gamma\mu\omega(\xi_i, \xi_j),\end{equation}
and the eigenvalue problem (\ref{eigenomega}) becomes: 
\begin{equation}\label{Aeing}A a = \tilde \lambda_k a.\end{equation}

This matrix can be considered as the equivalent of the  { \it affinity matrix} introduced 
 by Perona in \cite{Perona} to perform perceptual grouping. 
Perona proposed to model the affinity matrix in term of an euristic distance $d(\xi)$, facilitating collinear and cocircular couple of elements. 
Indeed by (\ref{omegadist}) we see that 
$$A_{ij} \simeq e^{-\it{d_c}^2(\xi_i, \xi_j) },$$
where $d_c$ is the Carnot Carath\'eodory distance defined in (\ref{dCC}).

\subsection{Spectral clustering and dimensionality reduction}
In \cite{Perona} the problem of perceptual grouping has been faced in terms of reduction of the complexity in the description of a scene. The visual scene is described in term of the affinity matrix $A_{ij}$ with a complexity of order $O(N^2)$  if N discrete elements are present in the scene. The idea of Perona and Freeman is to describe the scene approximating the matrix $A_{ij}$ by the sum of matrices of rank 1 and complexity $N$, each of which will identify a perceptual unit in the scene. 
If the number of the perceptual units present in the scene is much smaller than $N$, this procedure 
reduces the dimensionality of the description. 
A rank 1 matrix will be represented as the external product of a vector $p$ with itself. 

The first one will be computed as the best approximation of $A_{ij}$ minimizing the 
Frobenius norm as follows:
$$p_1= arg min_{\hat{p}} \sum_{i,j=1}^N(A_{ij}-\hat{p}_i \hat{p}_j)^2$$
where the term $p p^T=\sum_{i,j=1}^N \hat{p}_i \hat{p}_j $ is the rank one matrix with complexity order  $O(N)$.

Perona proved that the minimizer $p_1$ is the first eigenvector $v_1$ of the matrix $A$ with largest eigenvalue $\lambda_1$: $p_1=\lambda_1^{1/2}v_1.$

Then the problem is repeated on the vector space orthogonal to $p_1$. The minimizer will correspond to the second eigenvector, and iteratively the others eigenvectors are recovered. 
The process ends when the associated eigenvalue is sufficiently small. In this way in general only 
$n$ eigenvectors are selected, with $n<N$, leading to the dimensionality reduction.

Then the problem of grouping is reduced to the spectral analysis of the affinity matrix $A_{ij}$, where the salient objects in the scene correspond to the eigenvectors with largest eigenvalues.

We just showed in the previous paragraphs that this spectral analysis can be implemented 
by the neural population equation 
in the functional architecture of the primary visual cortex. 
We can now interpret eigenvectors of equation (\ref{Aeing}) as the gestalten 
segmenting the scene.

\section{Numerical simulation and results}

\subsection{Numerical approximation of the kernel}

\begin{figure}
\centering
\includegraphics[width=5cm]{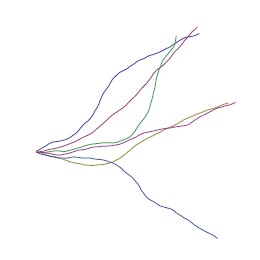}
\includegraphics[width=5cm]{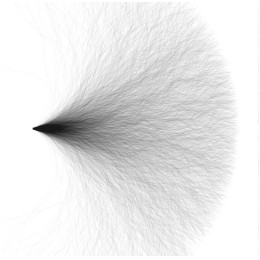}
\caption{Estimate of the fundamental solution $\Gamma$ of equation (\ref{gonzalo}) with the Markov Chain Monte Carlo method. It is visualized the projection of $\Gamma_d$ in the $(x,y)$ plane. 
On the left with $6$ random paths and on the right with $3000$ with $\sigma=0.08$, and $H=100$. }
\label{fig:path}
\end{figure}

\begin{figure}
\centering
\includegraphics[width=7cm]{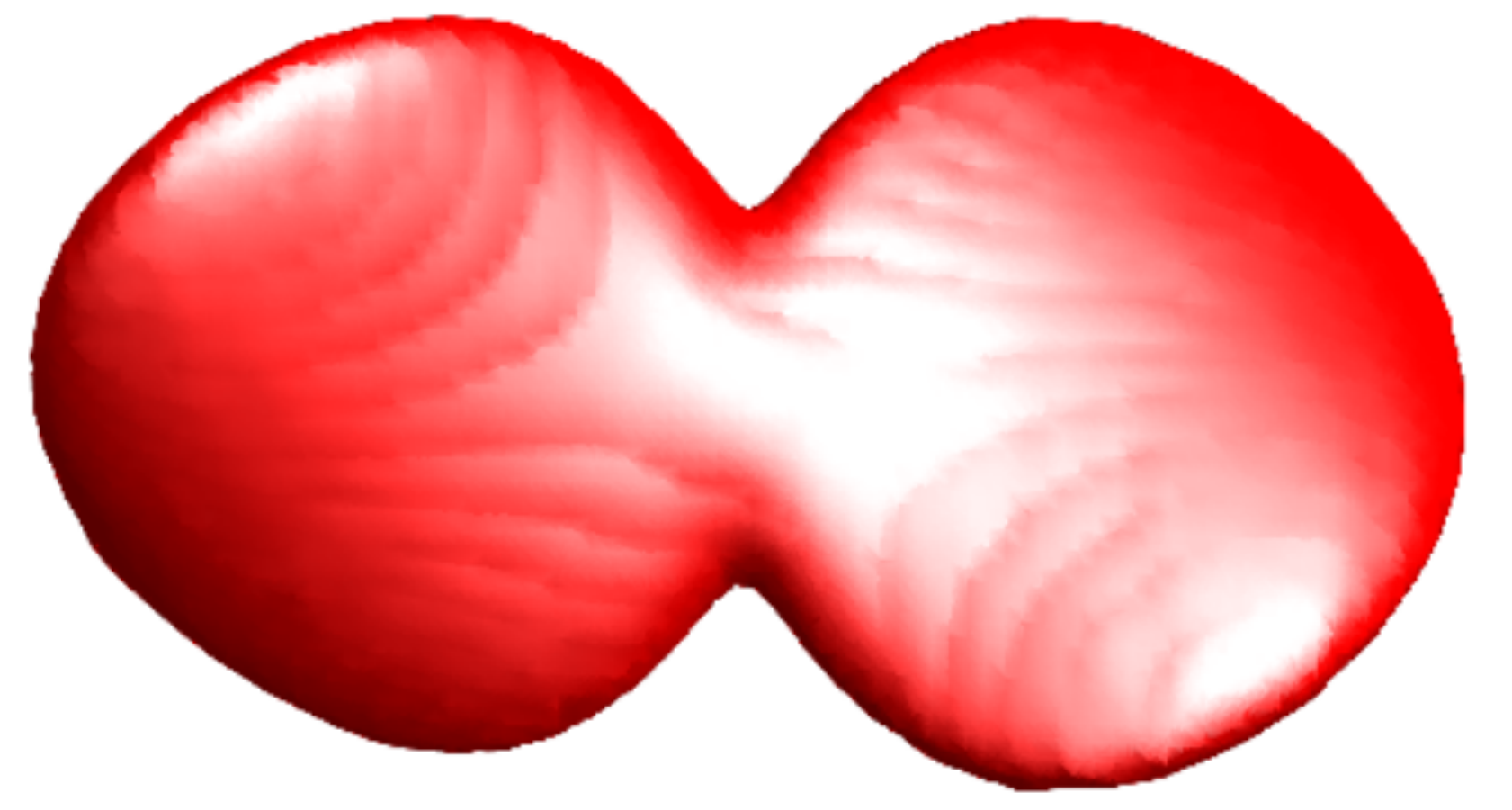}
\caption{A level set of the kernel $\omega$, obtained via the simmetrisation of the fundamental solution $\Gamma_d$}
 \label{fig44}
\end{figure}

We numerically evaluate the connectivity kernel $\omega$, defined by equation (\ref{omegakernel}).  
The fundamental solution $\Gamma$ of equation (\ref{gonzalo}) is numerically 
estimated with standard Markov Chain Monte Carlo methods
(MCMC) \cite{RobertCasella}. This is done by generating random paths obtained
from numerical solutions of the system (\ref{stoc}). This system is discretized as follows 

\begin{equation}\label{eq:discretePathV}
\left\{
\begin{array}{rcl}
x_{s+\Delta s} - x_{s} & = & \Delta s\cos(\theta) \vspace{4pt}\\
y_{s+\Delta s} - y_{s} & = & \Delta s\sin(\theta)\vspace{4pt}\\
\theta_{s+\Delta s} - \theta_{s} & = & \Delta s N(\sigma,0) \vspace{4pt}
\end{array}
\hspace{8pt},\hspace{8pt}s \in \{0,\dots,H\}
\right.
\end{equation}
where $H$ is the number of steps performed by the random path and $N(\sigma,0)$ is a generator of numbers taken from a normal distribution with mean $0$ and variance $\sigma$. Solving this finite difference $n$ times will give $n$ different realizations of the stochastic path: the estimated kernel  $\Gamma_d(\xi_i, \xi_0 )$,  is computed
averaging their passages over discrete volume elements, and
smoothing the results with local weighted means.
In Figure \ref{fig:path} a projection of the fundamental solution $\Gamma_d$ 
is visualized with different number of paths. 
In Figure \ref{fig44} a level set of the connectivity kernel $\omega$ is represented.
Since the connectivity is implemented in the 2-dimensional cortical layer we provide also a visualization of the kernel $\omega$ superimposed to the pinwheel structure.   
On the left of Figure a pinwheel structure $\tilde\theta(x; y)$ is visualized, outcome of a simulation following (BArbieri piwheel).  On the right the connectivity kernel is superimposed
to the pinwheels structure. Particularly the kernel is visualized by means of black points generated with a probability density proportional to the value of the kernel at the point $(x, y, \tilde\theta(x, y))$. The comparison of the image with the results of Bosking presented in Figure \ref{tracer} 
shows that the kernel $\omega$ provides a good  estimate of the measured cortical connectivity. 

\begin{figure}
\centering
\includegraphics[height=4.5cm]{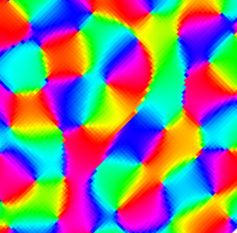}
\includegraphics[height=4.5cm]{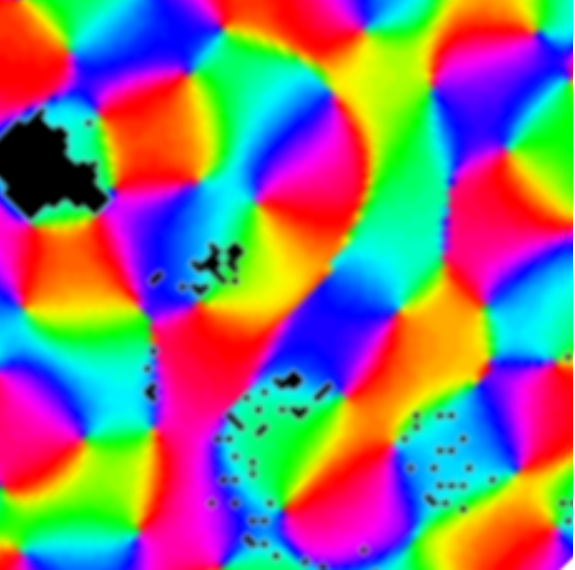}
\caption{On the left the pinwheel structure $\tilde \theta(x,y)$ is visualized, simulated as in \cite{}. On the right the connectivity kernel $\omega$ is superimposed to the pinwheels structure. The density of black points is proportional to the value of $\omega(x,y, \tilde \theta(x,y))$.}\label{simulatepin}
\end{figure}

\subsection{Results of grouping}
\begin{figure}
\centering
\includegraphics[width=5.5cm]{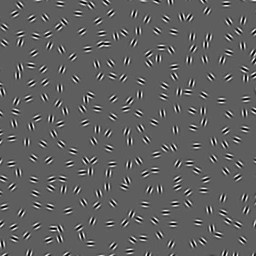}
\includegraphics[width=5.5cm]{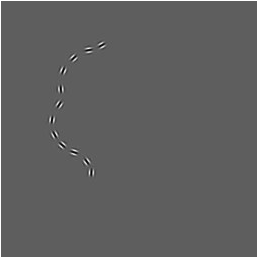}
\caption{The experiment of Fields, Heyes and Hess. The proposed stimulus (on the left) and the perceptual unit present in it (right) \cite{Field1993}
 }\label{fig33}
\end{figure}

In \cite{Field1993} Fields, Heyes and Hess experimented the ability of the human visual system to detect perceptual units out of a random distribution of oriented elements. 
 In Figure \ref{fig33} (left) it is shown the stimulus proposed to the observer, 
from which the visual system is able to individuate the perceptual unit shown in the right. In the following we will test our grouping model on similar stimuli to individuate the perceptual units present in the images.

\begin{figure}
\centering
\includegraphics[height=4.5cm]{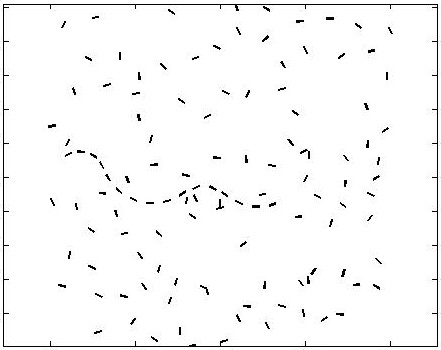}
\includegraphics[height=4.5cm]{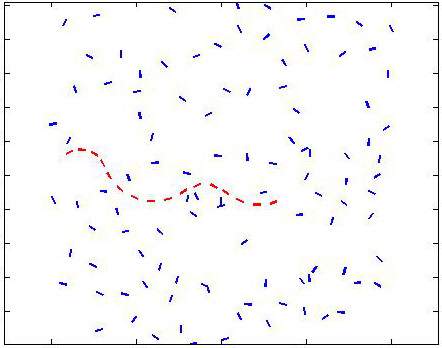}
\caption{In the image on the left a random distribution of segments and a coherent structure are present. On the right the first eigenvector of the affinity matrix is shown. In red are visualized the segments on which the eigenvector is greater than a given threshold.}\label{fig34}
\end{figure}

In the first experiment we considered 150 position-orientation patches, with coordinates $\xi_i$. A subset of elements is organized in a coherent way and the large majority is randomly chosen, in a way similar to the experiment of \cite{Field1993}(see Figure \ref{fig34}, left). 
These points define a domain $\Omega_d=\{\xi_i: i=1, \cdots n\}$ as 
in equation (\ref{omegadiscrete}), and we will define the input stimulus $h$ as a function defined on the whole cortical space $M$, which attains value $c$ on $\Omega$ and $0$ outside. 

\begin{figure}
\centering
\includegraphics[height=5.5cm]{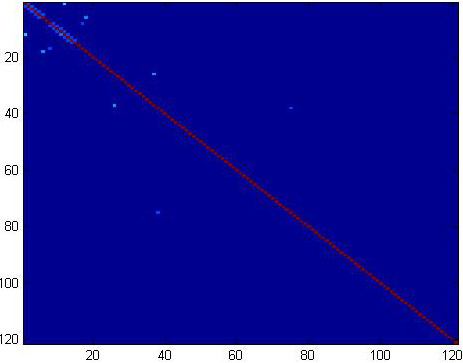}
\includegraphics[height=5.5cm]{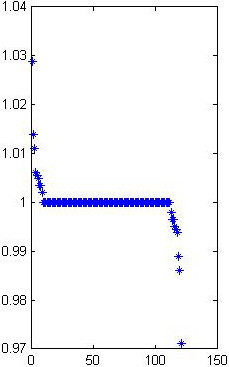}
\caption{On the left is visualized the affinity matrix. On the right its eigenvalues are shown.}\label{fig35}
\end{figure}

The connectivity among these elements is defined as in equation (\ref{affinemat}), by means of the connectivity kernel $\gamma\mu\omega(\xi_i, \xi_j)$. The entries of the associated matrix $A_{ij}$ are visualized in Figure  \ref{fig35}. It is evident the quasi block structure of the matrix with a principal block on the top left and small blocks on the 
quasidiagonal structure. The principal block corresponds to the coherent object and the diagonal to the correlated ones. The eigenvalue problem  (\ref{Aeing}) is faced and eigenvalues of the associated affinity matrix are computed. 

  Figure  \ref{fig35} right shows the ordered distributions of eigenvalues, where a dominant eigenvalue is present. The corresponding eigenvector is visualized in Figure \ref{fig34}
(right) and individuates the coherent perceptual unit. 

\begin{figure}
\includegraphics[height=4.5cm]{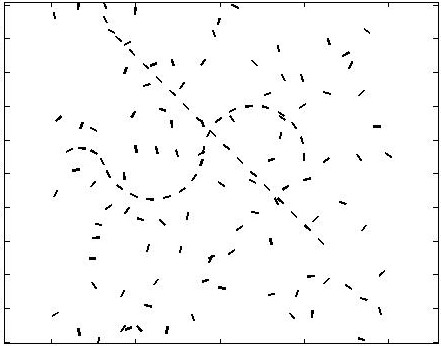}
\includegraphics[height=4.5cm]{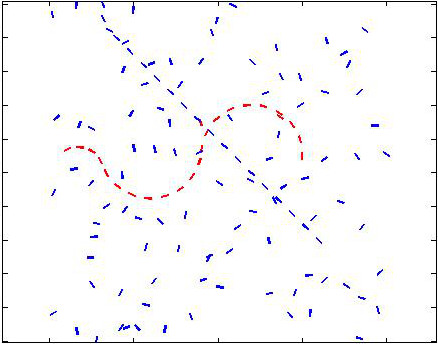}\\
\includegraphics[height=4.5cm]{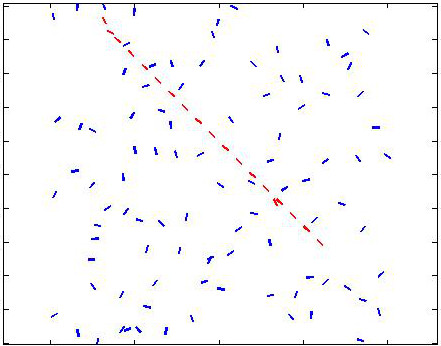}
\includegraphics[height=4.5cm]{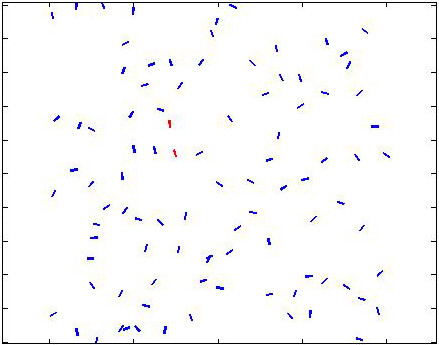}
\caption{A stimulus containing 2 perceptual units (top left)  is segmented. After that the first eigenvector of the affinity matrix is computed (top right), the affinity matrix is updated removing the detected perceptual unit. The first eigenvector of the updated affinity matrix is visualized (bottom left). The procedure is iterated for the next unit (bottom right).}\label{fig35percept}
\end{figure}

In the second experiment a stimulus containing 2 perceptual units is present. As before we compute the connectivity kernel $\gamma\mu\omega(\xi_i, \xi_j)$ and the associated matrix $A_{ij}$. The eigenvalue problem  (\ref{Aeing}) is faced and eigenvalues of the associated affinity matrix are computed. 
The first eigenvector of the affinity matrix is computed and shown in Figure \ref{fig35percept} (top right). After that the affinity matrix is updated removing the detected perceptual unit. The first eigenvector of the updated affinity matrix is visualized in Figure \ref{fig35percept} (bottom left). The procedure is iterated for the next unit which only contains two oriented element (bottom right).

\end{document}